\documentclass{article}

     \PassOptionsToPackage{numbers, compress}{natbib}

    \usepackage[final]{neurips_2022}

\usepackage[ruled,vlined]{algorithm2e}
\usepackage{algorithmic}
\usepackage[utf8]{inputenc} 
\usepackage[T1]{fontenc}    
\usepackage{hyperref}       
\usepackage{url}            
\usepackage{booktabs}       
\usepackage{amsfonts}       
\usepackage{nicefrac}       
\usepackage{microtype}      
\usepackage{xcolor}         
\usepackage{graphicx}
\usepackage{subfigure}
\usepackage{makecell}
\graphicspath{ {./figures/} }

\title{Fast DistilBERT on CPUs}

\author{%
Haihao Shen$^*$ \\
Intel Corporation \\
\texttt{haihao.shen@intel.com}
\And
Ofir Zafrir$^*$ \\
Intel Labs \\
\texttt{ofir.zafrir@intel.com}
\And
Bo Dong \\
Intel Corporation \\
\texttt{bo1.dong@intel.com}
\And
Hengyu Meng \\
Intel Corporation \\
\texttt{hengyu.meng@intel.com}
\And
Xinyu Ye \\
Intel Corporation \\
\texttt{xinyu.ye@intel.com}
\And
Zhe Wang \\
Intel Corporation \\
\texttt{zhe1.wang@intel.com}
\And
Yi Ding \\
Intel Corporation \\
\texttt{yi1.ding@intel.com}
\And
Hanwen Chang \\
Intel Corporation \\
\texttt{hanwen.chang@intel.com}
\And
Guy Boudoukh \\
Intel Labs \\
\texttt{guy.boudoukh@intel.com}
\And
Moshe Wasserblat \\
Intel Labs \\
\texttt{moshe.wasserblat@intel.com}
}

\begin{document}

\maketitle
\def\thefootnote{*}\footnotetext{Equal contribution}
\begin{abstract}
Transformer-based language models have become the standard approach to solving natural language processing tasks.
However, industry adoption usually requires the maximum throughput to comply with certain latency constraints that prevents Transformer models from being used in production.
To address this gap, model compression techniques such as quantization and pruning may be used to improve inference efficiency.
However, these compression techniques require specialized software to apply and deploy at scale.
In this work, we propose a new pipeline for creating and running Fast Transformer models on CPUs, utilizing hardware-aware pruning, knowledge distillation, quantization, and our own Transformer inference runtime engine with optimized kernels for sparse and quantized operators.
We demonstrate the efficiency of our pipeline by creating a Fast DistilBERT model showing minimal accuracy loss on the question-answering SQuADv1.1 benchmark, and throughput results under typical production constraints and environments.
Our results outperform existing state-of-the-art Neural Magic's DeepSparse runtime performance by up to 50\% and up to 4.1x performance speedup over ONNX Runtime. Source code is publicly available at https://github.com/intel/intel-extension-for-transformers.
\end{abstract}

\section{Introduction and related work}
\label{sec:intro}
Large Transformer-based Language Models (LMs) are evolving rapidly from millions of parameters, e.g., BERT-Large~\cite{devlin2018bert}, to billions of parameters, e.g., Turing-Megatron~\cite{smith2022using}, and GPT3~\cite{Brown2020LanguageMA}.
Those large models have demonstrated promising state-of-the-art (SoTA) accuracy on a wide range of NLP tasks.
However, those models are inefficient and therefore unsuited to the limited computational sources and strict latency constraints in production.

To enable and increase the efficiency and scale of deployed Transformer models, additional model compression and optimization are usually required along with a dedicated inference engine.
For example, DistilBERT~\cite{sanh2019distilbert}, using knowledge distillation~\cite{Hinton2015DistillingTK}, shows minimal impact on downstream NLP tasks at a reduced size by 40\%.
Additionally, pruning~\cite{sanh2020movement} and quantization~\cite{zafrir2019q8bert} are well-known techniques to further compress a Transformer model.
Recent works have proposed pruning Transformer models at pre-training to create sparse pre-trained LMs.
It has been established that fine-tuning these sparse pre-trained LMs to downstream tasks can produce results that are competitive with other pruning methods, obviating the need for pruning.~\cite{zafrir2021prune, kurtic2022optimal}.
For example, a DistilBERT with 90\% unstructured sparsity was produced with minimal impact on the accuracy of downstream tasks~\cite{zafrir2021prune}; however, performance gains have not been demonstrated for this model.
Similarly, Neural Magic published the throughput performance of a structured sparse DistilBERT with 80\% sparsity, batch size 32 and sequence length 128, although latency performance was not provided~\cite{neuralmagic}.
Nvidia recently proposed novel 2:4 structured sparsity that is only available on Ampere architecture and above~\cite{pool2021accelerating}.
Quantization~\cite{jacob2018quantization}\cite{wang2019haq} has been maturing in industry with two well-known approaches: Post-Training Quantization (PTQ) and Quantization-Aware Training (QAT).
Typically, PTQ requires an offline calibration process on a representative calibration dataset, while QAT requires an additional fine-tuning stage simulating quantization inference while training.
Although the approaches have proved successful in some typical models~\cite{yao2022zeroquant,zafrir2019q8bert}, it remains challenging to produce a quantized model based on a highly sparse model without sacrificing accuracy.

In this paper, we propose a new pipeline for creating and running Fast Transformer models on CPUs.
We extend the model compression approach PruneOFA~\cite{zafrir2021prune} by enabling CPU-friendly block-wise structured sparsity.
We then apply an accuracy-aware post-training quantization approach to generate an 8-bit (INT8) sparse model.
Furthermore, we demonstrate our models using our Transformer inference engine dedicated for running sparse \& quantized Transformers on CPUs.
We apply this pipeline on DistilBERT~\cite{sanh2019distilbert} to create Fast DistilBERT.
We show that Fast DistilBERT can meet a requirement of under 1\% accuracy loss vs. the DistilBERT baseline on the question-answering benchmark SQuADv1.1~\cite{rajpurkar2016squad}.
We demonstrate up to 1.5x performance gain over Neural Magic's DeepSparse engine~\cite{pmlr-v119-kurtz20a} and up to 4.1x performance gain over ONNX Runtime on common CPUs from Amazon Web Services (AWS) under typical production constraints.

Our main contributions are threefold:
1) Propose a hardware-aware extreme compression technique for fast Transformer models on CPUs.
2) Create an efficient Transformer inference runtime for sparse \& quantized Transformer models.
3) Demonstrate new SOTA performance under typical constraints in common production environments.

\section{Method description}
\label{sec:compression}
In this section, we describe how to solve the challenges of applying model compression techniques on Transformer-based LMs.

\paragraph{Block-wise structured sparsity}
\label{sec:sparsity}
In this work, we accelerate sparse Transformer-based models with specialized sparse GEMM operators which require a structured sparsity of constant size blocks in the output dimension, see Section~\ref{sec:sparse-gemm-ops}.
To that end, we extend the model compression infrastructure proposed by \citet{zafrir2021prune} to create sparse pre-trained LMs with block-wise structured sparsity.
Further details on block-wise pruning in Appendix~\ref{sec:app_block_pruning}.

\paragraph{Fine-tuning with knowledge distillation}
\label{sec:fine-tuning}
Knowledge distillation is a widely used method for model compression \cite{Hinton2015DistillingTK,sanh2019distilbert}.
Knowledge distillation while fine-tuning Transformer-based models may improve the accuracy of the model on the downstream task and bridge the accuracy gap caused by compression methods applied to the model \cite{sanh2020movement,zafrir2021prune,kurtic2022optimal}.
Following these works we apply knowledge distillation while fine-tuning the block-wise sparse pre-trained LM to downstream tasks to mitigate the accuracy loss.

\paragraph{Post-training quantization}
\label{sec:quantization}
PTQ is an effective approach to quantizing a model without additional training steps.
It requires a calibration process using a representative dataset to determine the quantization parameters of the model.
Converting FP32 operations to INT8 can increase their efficiency by up to 4x \cite{bhandare2019efficient}.
We apply PTQ with automatic accuracy-aware tuning on the sparse Transformer model to produce an optimized model using Intel\textsuperscript{\tiny\textregistered} Neural Compressor (INC) open-source tool for quantization \cite{inc}.

\section{Software acceleration}
\label{sec:software}
We develop a Transformer inference engine on CPU with advanced runtime, graph optimization, and sparse GEMM operators.

\paragraph{Advanced runtime}
We develop an advanced, cache friendly, memory allocator to enable buffer reuse and to reduce memory allocation overhead, in particular for a continuous demand to allocate/free small memory chunks.
Moreover, we implement weight sharing that allows a single copy of weight to be shared across multiple instances running in parallel during inference.
More details in Appendix~\ref{sec:advanced_allocator},\ref{sec:wsharing}.

\paragraph{Graph optimization}
The computation graph of a Transformer-based model contains many small and/or redundant operations.
After obtaining our quantized Transformer-based model we apply several optimizations at the graph level including operations fusion, removal, etc. See Appendix~\ref{sec:gopts} for further details.

\begin{figure}[htbp]
    \centering
    \includegraphics[scale=0.55]{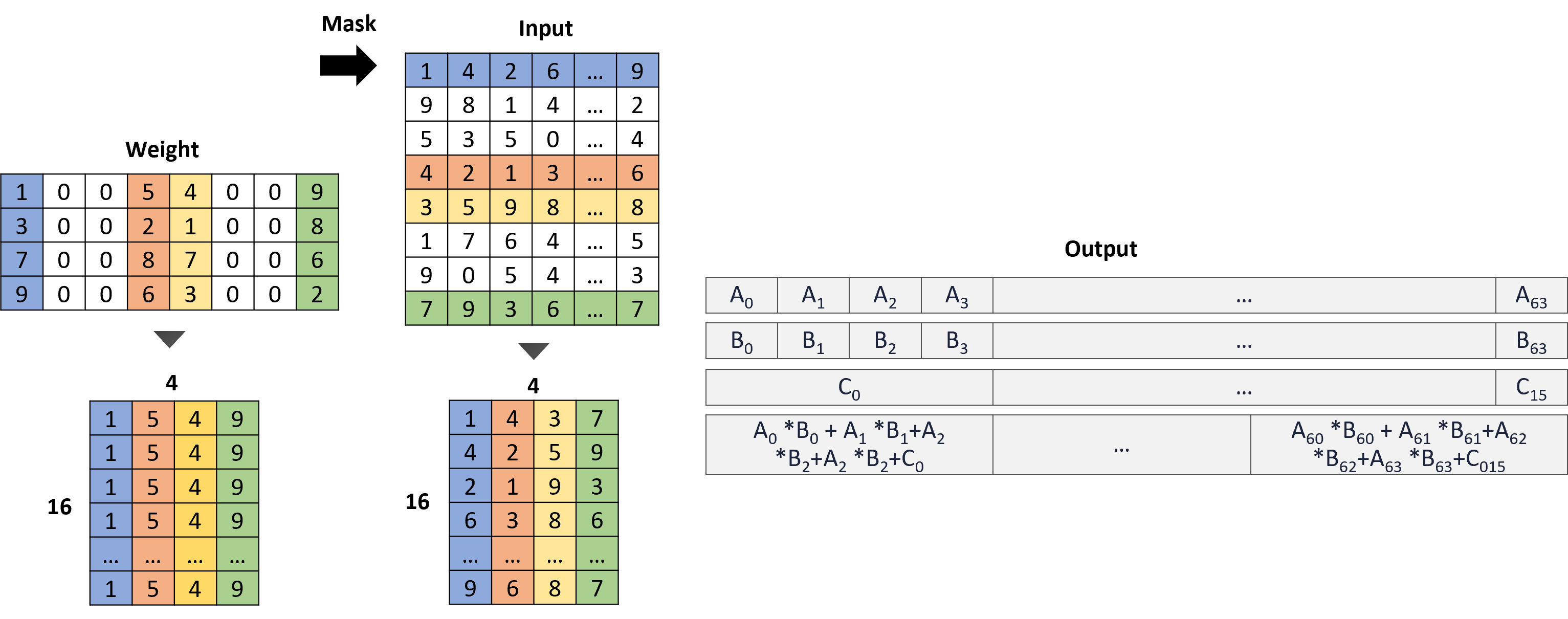}
    \caption{INT8 sparse GEMM kernel based on VNNI}
    \label{fig:sparse}
\end{figure}

\paragraph{Sparse GEMM operators}
\label{sec:sparse-gemm-ops}
We implement INT8 sparse GEMM kernel to accelerate operations between dense input and sparse weights leveraging AVX512-VNNI instructions.
Figure~\ref{fig:sparse} illustrates how the kernel operates.
Given a block-wise sparse weight, we broadcast the non-zero weight block to form a VNNI-format block A.
Based on the mask in the sparse weight, we re-organize the corresponding input as another VNNI-format block B. Then, the kernel uses VNNI to produce the intermediate output given A and B, and add bias C as the final output.
Algorithm~\ref{algo:sparse} in the appendix contains the implementation of the kernel.

\section{Experimental setup}
\label{sec:setup}
To demonstrate the performance of the runtime optimizations described in Section~\ref{sec:software}, we produce a block-wise sparse pre-trained DistilBERT, and then fine-tune and quantize it for the question-answering SQuADv1.1 benchmark \cite{rajpurkar2016squad} using the methods described in Section~\ref{sec:compression}.
Appendix~\ref{sec:app_model_preparation} details our model creation process.
Table~\ref{tab:models} lists the F1 scores of the DistilBERT baseline and our compressed model.
To evaluate the performance of our optimizations, we evaluate our compressed model performance on a AWS c6i.12xlarge CPU instance for 3 scenarios: maximum ThroughPut (TP), minimum latency, and production.
The production scenario requires maximum TP under 10 milliseconds (ms) latency per batch.
Moreover, we compare our performance to other public inference runtimes for quantized and sparse models, ONNX Runtime and Neural Magic.
We include the F1 results for their corresponding models in Table~\ref{tab:models}.

\begin{table}[htbp]
\caption{DistilBERT baseline and compressed models -- SQuADv1.1 results}
\label{tab:models}
\centering
\begin{tabular}{lcl}
               \toprule
               Model & F1 Score & Model Source/Generation\\
               \midrule
               FP32 dense (baseline) & 85.80\% & Taken from~\cite{sanh2019distilbert}\\
               INT8 sparse (ONNX Runtime) & 85.64\% & Generated by INC~\cite{inc} \\
               INT8 sparse (Neural Magic) & 86.26\% & Downloaded from ~\cite{neuralmagic} \\
               INT8 sparse (Ours) & 86.07\% & Generated by INC~\cite{inc} \\
               \bottomrule
\end{tabular}
\end{table}

\section{Results}
\label{sec:results}
\begin{figure}[htbp]
\centering
    \includegraphics[scale=0.65]{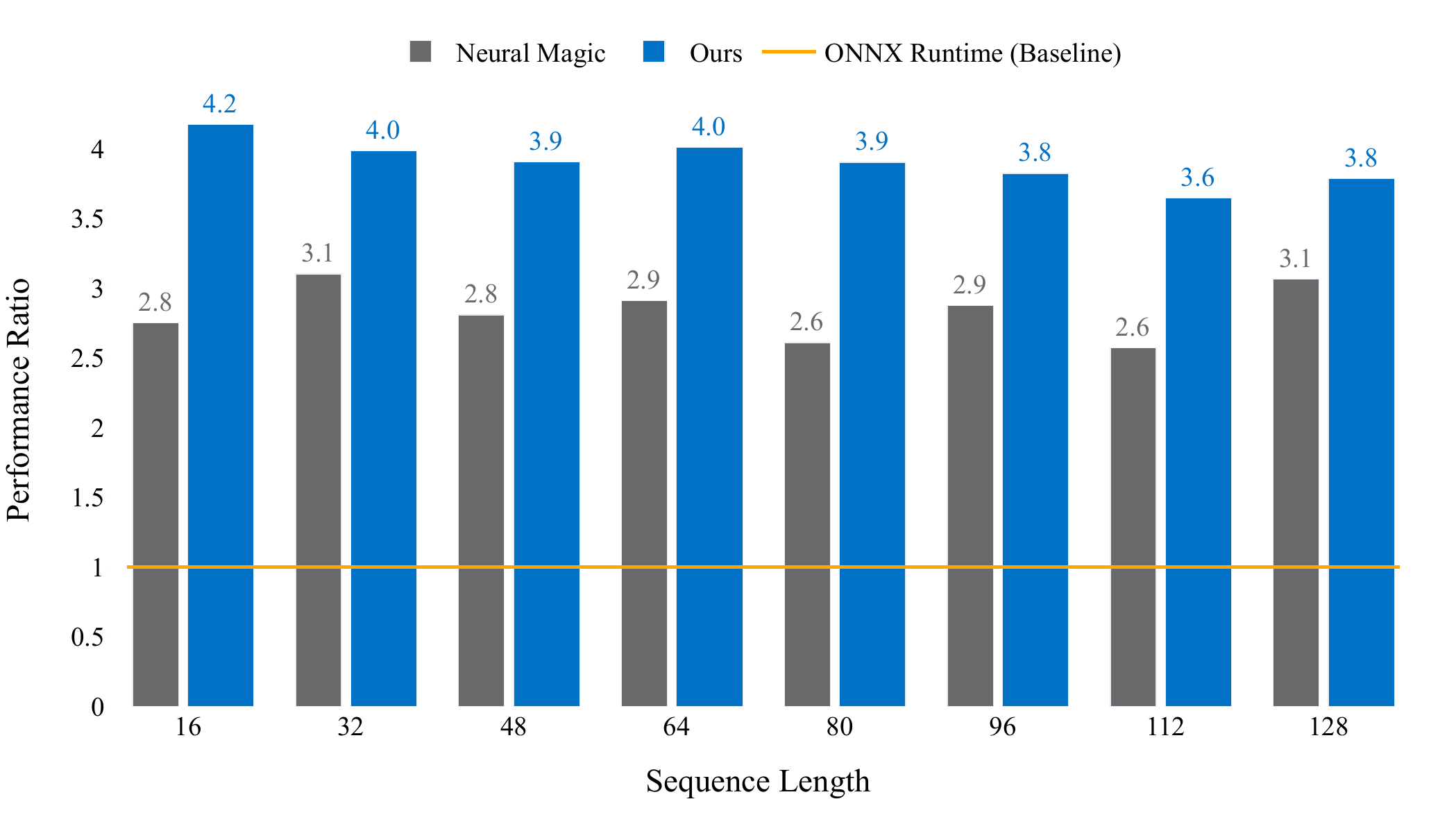}
    \caption{Production scenario results, maximum TP under 10ms inference latency}
    \label{fig:perf}
\end{figure}

All the results we present are an average over several measurements after several warm-up iterations.
Figure~\ref{fig:perf} presents the relative performance of DistilBERT compressed models for the production scenario.
Table~\ref{throughput_under_latency} in Appendix~\ref{sec:model_details} shows the absolute performance accordingly.
Our solution shows a consistent performance gain of 3.6x-4.1x over ONNX Runtime and outperforms Neural Magic by up to 50\% for a range of sequence lengths.
Table~\ref{throughput_latency}) presents the results for the maximum TP and minimum latency scenarios.
We observe that our solution yields results comparable to the Neural Magic solution and performs 2x-3x better than ONNX Runtime.
Our Transformer inference solution demonstrates high efficiency on common CPU production environments.
To the best of our knowledge, we offer the best inference solution for Transformer deployment on CPU.

\begin{table}[htbp]
\caption{Maximum TP and minimum latency results with sequence length 32}
\label{throughput_latency}
\centering
\begin{tabular}{lcc}
              \toprule
              Model & \makecell{Maximum Throughput\\(Samples/second)} & \makecell{Minimal Latency\\(Milliseconds/sample)} \\
              \midrule
              INT8 sparse (ONNX Runtime) & 1920 & 2.76 \\
              INT8 sparse (Neural Magic) & 5904 & 1.24 \\
              INT8 sparse (Ours) & 6002 & 1.27 \\
              \bottomrule
\end{tabular}
\end{table}
\section{Summary and future work}
\label{sec:summary}
In this paper, we presented a combined hardware-aware model compression technique (block-wise structured sparsity, knowledge distillation, and quantization) and demonstrated it with DistilBERT.
We created a new dedicated inference engine which unlocks the performance of extremely compressed Transformer-based LMs on CPUs.
Our results outperform Neural Magic's inference solution by up to 50\% and demonstrate up to 4.1x better performance than ONNX Runtime under production constraints.
We plan to apply the new model compression technique to other popular Transformer-based models and demonstrate the inference efficiency using our inference solution.
Code and models for reproducing the published results are available at https://github.com/intel/intel-extension-for-transformers.

\bibliographystyle{abbrvnat}
\bibliography{references}

\begin{thebibliography}{19}
\providecommand{\natexlab}[1]{#1}
\providecommand{\url}[1]{\texttt{#1}}
\expandafter\ifx\csname urlstyle\endcsname\relax
  \providecommand{\doi}[1]{doi: #1}\else
  \providecommand{\doi}{doi: \begingroup \urlstyle{rm}\Url}\fi

\bibitem[Bhandare et~al.(2019)Bhandare, Sripathi, Karkada, Menon, Choi, Datta,
  and Saletore]{bhandare2019efficient}
A.~Bhandare, V.~Sripathi, D.~Karkada, V.~Menon, S.~Choi, K.~Datta, and
  V.~Saletore.
\newblock Efficient 8-bit quantization of transformer neural machine language
  translation model.
\newblock \emph{arXiv preprint arXiv:1906.00532}, 2019.

\bibitem[Brown et~al.(2020)Brown, Mann, Ryder, Subbiah, Kaplan, Dhariwal,
  Neelakantan, Shyam, Sastry, Askell, Agarwal, Herbert-Voss, Krueger, Henighan,
  Child, Ramesh, Ziegler, Wu, Winter, Hesse, Chen, Sigler, Litwin, Gray, Chess,
  Clark, Berner, McCandlish, Radford, Sutskever, and
  Amodei]{Brown2020LanguageMA}
T.~B. Brown, B.~Mann, N.~Ryder, M.~Subbiah, J.~Kaplan, P.~Dhariwal,
  A.~Neelakantan, P.~Shyam, G.~Sastry, A.~Askell, S.~Agarwal, A.~Herbert-Voss,
  G.~Krueger, T.~Henighan, R.~Child, A.~Ramesh, D.~M. Ziegler, J.~Wu,
  C.~Winter, C.~Hesse, M.~Chen, E.~Sigler, M.~Litwin, S.~Gray, B.~Chess,
  J.~Clark, C.~Berner, S.~McCandlish, A.~Radford, I.~Sutskever, and D.~Amodei.
\newblock Language models are few-shot learners.
\newblock \emph{ArXiv}, abs/2005.14165, 2020.

\bibitem[Devlin et~al.(2018)Devlin, Chang, Lee, and Toutanova]{devlin2018bert}
J.~Devlin, M.-W. Chang, K.~Lee, and K.~Toutanova.
\newblock Bert: Pre-training of deep bidirectional transformers for language
  understanding.
\newblock \emph{arXiv preprint arXiv:1810.04805}, 2018.

\bibitem[Hinton et~al.(2015)Hinton, Vinyals, and Dean]{Hinton2015DistillingTK}
G.~E. Hinton, O.~Vinyals, and J.~Dean.
\newblock Distilling the knowledge in a neural network.
\newblock \emph{ArXiv}, abs/1503.02531, 2015.

\bibitem[Intel(2020)]{inc}
Intel.
\newblock Neural compressor.
\newblock \emph{https://github.com/intel/neural-compressor}, 2020.

\bibitem[Jacob et~al.(2018)Jacob, Kligys, Chen, Zhu, Tang, Howard, Adam, and
  Kalenichenko]{jacob2018quantization}
B.~Jacob, S.~Kligys, B.~Chen, M.~Zhu, M.~Tang, A.~Howard, H.~Adam, and
  D.~Kalenichenko.
\newblock Quantization and training of neural networks for efficient
  integer-arithmetic-only inference.
\newblock In \emph{Proceedings of the IEEE conference on computer vision and
  pattern recognition}, pages 2704--2713, 2018.

\bibitem[Kurtic et~al.(2022)Kurtic, Campos, Nguyen, Frantar, Kurtz, Fineran,
  Goin, and Alistarh]{kurtic2022optimal}
E.~Kurtic, D.~Campos, T.~Nguyen, E.~Frantar, M.~Kurtz, B.~Fineran, M.~Goin, and
  D.~Alistarh.
\newblock The optimal bert surgeon: Scalable and accurate second-order pruning
  for large language models.
\newblock \emph{arXiv preprint arXiv:2203.07259}, 2022.

\bibitem[Kurtz et~al.(2020)Kurtz, Kopinsky, Gelashvili, Matveev, Carr, Goin,
  Leiserson, Moore, Nell, Shavit, and Alistarh]{pmlr-v119-kurtz20a}
M.~Kurtz, J.~Kopinsky, R.~Gelashvili, A.~Matveev, J.~Carr, M.~Goin,
  W.~Leiserson, S.~Moore, B.~Nell, N.~Shavit, and D.~Alistarh.
\newblock Inducing and exploiting activation sparsity for fast inference on
  deep neural networks.
\newblock In H.~D. III and A.~Singh, editors, \emph{Proceedings of the 37th
  International Conference on Machine Learning}, volume 119 of
  \emph{Proceedings of Machine Learning Research}, pages 5533--5543, Virtual,
  13--18 Jul 2020. PMLR.
\newblock URL \url{http://proceedings.mlr.press/v119/kurtz20a.html}.

\bibitem[NeuralMagic(2022)]{neuralmagic}
NeuralMagic.
\newblock Sparsezoo.
\newblock \emph{https://sparsezoo.neuralmagic.com/}, 2022.

\bibitem[Pool et~al.(2021)Pool, Sawarkar, and Rodge]{pool2021accelerating}
J.~Pool, A.~Sawarkar, and J.~Rodge.
\newblock Accelerating inference with sparsity using the nvidia ampere
  architecture and nvidia tensorrt.
\newblock \emph{NVIDIA Developer Blog}, 2021.

\bibitem[Rajpurkar et~al.(2016)Rajpurkar, Zhang, Lopyrev, and
  Liang]{rajpurkar2016squad}
P.~Rajpurkar, J.~Zhang, K.~Lopyrev, and P.~Liang.
\newblock Squad: 100,000+ questions for machine comprehension of text.
\newblock \emph{arXiv preprint arXiv:1606.05250}, 2016.

\bibitem[Sanh et~al.(2019)Sanh, Debut, Chaumond, and Wolf]{sanh2019distilbert}
V.~Sanh, L.~Debut, J.~Chaumond, and T.~Wolf.
\newblock Distilbert, a distilled version of bert: smaller, faster, cheaper and
  lighter.
\newblock \emph{arXiv preprint arXiv:1910.01108}, 2019.

\bibitem[Sanh et~al.(2020)Sanh, Wolf, and Rush]{sanh2020movement}
V.~Sanh, T.~Wolf, and A.~Rush.
\newblock Movement pruning: Adaptive sparsity by fine-tuning.
\newblock \emph{Advances in Neural Information Processing Systems},
  33:\penalty0 20378--20389, 2020.

\bibitem[Smith et~al.(2022)Smith, Patwary, Norick, LeGresley, Rajbhandari,
  Casper, Liu, Prabhumoye, Zerveas, Korthikanti, et~al.]{smith2022using}
S.~Smith, M.~Patwary, B.~Norick, P.~LeGresley, S.~Rajbhandari, J.~Casper,
  Z.~Liu, S.~Prabhumoye, G.~Zerveas, V.~Korthikanti, et~al.
\newblock Using deepspeed and megatron to train megatron-turing nlg 530b, a
  large-scale generative language model.
\newblock \emph{arXiv preprint arXiv:2201.11990}, 2022.

\bibitem[Wang et~al.(2019)Wang, Liu, Lin, Lin, and Han]{wang2019haq}
K.~Wang, Z.~Liu, Y.~Lin, J.~Lin, and S.~Han.
\newblock Haq: Hardware-aware automated quantization with mixed precision.
\newblock In \emph{Proceedings of the IEEE/CVF Conference on Computer Vision
  and Pattern Recognition}, pages 8612--8620, 2019.

\bibitem[Yao et~al.(2022)Yao, Aminabadi, Zhang, Wu, Li, and
  He]{yao2022zeroquant}
Z.~Yao, R.~Y. Aminabadi, M.~Zhang, X.~Wu, C.~Li, and Y.~He.
\newblock Zeroquant: Efficient and affordable post-training quantization for
  large-scale transformers.
\newblock \emph{arXiv preprint arXiv:2206.01861}, 2022.

\bibitem[Zafrir et~al.(2019)Zafrir, Boudoukh, Izsak, and
  Wasserblat]{zafrir2019q8bert}
O.~Zafrir, G.~Boudoukh, P.~Izsak, and M.~Wasserblat.
\newblock Q8bert: Quantized 8bit bert.
\newblock In \emph{2019 Fifth Workshop on Energy Efficient Machine Learning and
  Cognitive Computing-NeurIPS Edition (EMC2-NIPS)}, pages 36--39. IEEE, 2019.

\bibitem[Zafrir et~al.(2021)Zafrir, Larey, Boudoukh, Shen, and
  Wasserblat]{zafrir2021prune}
O.~Zafrir, A.~Larey, G.~Boudoukh, H.~Shen, and M.~Wasserblat.
\newblock Prune once for all: Sparse pre-trained language models.
\newblock \emph{arXiv preprint arXiv:2111.05754}, 2021.

\bibitem[Zhu et~al.(2015)Zhu, Kiros, Zemel, Salakhutdinov, Urtasun, Torralba,
  and Fidler]{zhu2015aligning}
Y.~Zhu, R.~Kiros, R.~Zemel, R.~Salakhutdinov, R.~Urtasun, A.~Torralba, and
  S.~Fidler.
\newblock Aligning books and movies: Towards story-like visual explanations by
  watching movies and reading books.
\newblock In \emph{Proceedings of the IEEE international conference on computer
  vision}, pages 19--27, 2015.

\end{thebibliography}

\appendix

\section{Appendix}
Due to the space limitation of the main , we provide the additional details on model compression, software acceleration, and performance measurement in this appendix.

\subsection{Block-wise structured sparsity}
\label{sec:app_block_pruning}
The required sparsity pattern by our sparse GEMM operators, described in Section~\ref{sec:sparse-gemm-ops}, is 1-dimension blocks of size 4 in the output dimension of the weight in the case of a Linear layer.
We extend the pruning framework proposed by \citet{zafrir2021prune}.
In each pruning step a score is computed for each block with a pre-defined heuristic.
The blocks with the lowest scores' magnitude are pruned to reach the required sparsity ratio.
In this work, the use the average heuristic.
The score for each block is the average of the magnitudes of the weights inside the block.

\subsection{Model preparation}
\label{sec:app_model_preparation}
Following the PruneOFA~\cite{zafrir2021prune} process we create a sparse pre-trained LM with the required block-wise sparsity as described in Section~\ref{sec:sparsity}.
We take English Wikipedia and BookCorpus~\cite{zhu2015aligning} as our pre-training dataset.
We fine-tune BERT-Base\footnote{\url{https://huggingface.co/bert-base-uncased}} on the pre-training dataset in the teacher preparation step.
In the student pruning step we prune DistilBERT\footnote{\url{https://huggingface.co/distilbert-base-uncased}} in the required block-wise structured pattern with knowledge distillation from the teacher we prepared in the previous step.

After obtaining the pre-trained sparse DistilBERT we fine-tune it with pattern-lock\cite{zafrir2021prune} to SQuADv1.1 question-answering benchmark using the following  hyper-parameters in
Table~\ref{tb:hp}, where $\lambda_{kd}$ denotes student-teacher paired loss and $\lambda_{MLM}$ denotes student-ground truth loss.

\begin{table}[htbp]
 \caption{Hyper-parameters for sparse DistilBERT}
 \label{tb:hp}
\centering
\begin{tabular}{lc}
               \toprule
               Hyper-parameter & Value \\
               \midrule
               Learning rate & 0.00018 \\
               Batch Size & 12 \\
               Weight decay & 0.01 \\
               Epochs & 8 \\
               Learning rate decay & Linear \\
               Warmup ratio & 0.05 \\
               Sequence length & 384 \\
               $\lambda_{MLM}$ & 0 \\
               $\lambda_{kd}$ & 1 \\
               Temperature & 2 \\
               \bottomrule
 \end{tabular}
\end{table}

Note that we successfully demonstrate an end-to-end CPU solution on AWS from fine-tuning with above hyper-parameters on a c6i.32xlarge instance to inference on a c6i.12xlarge instance without any special hardware like GPUs or purpose-built accelerators.

\subsection{Software acceleration}
\label{sec:app_sw_acceleration}
In this section, we describe the advanced memory allocator, weight sharing, graph optimization, and sparse GEMM operators which are required to build up our inference solution and demonstrate the high inference efficiency.

\subsubsection{Advanced memory allocator}
\label{sec:advanced_allocator}
\begin{figure}[htbp]
    \centering
    \includegraphics[scale=0.4]{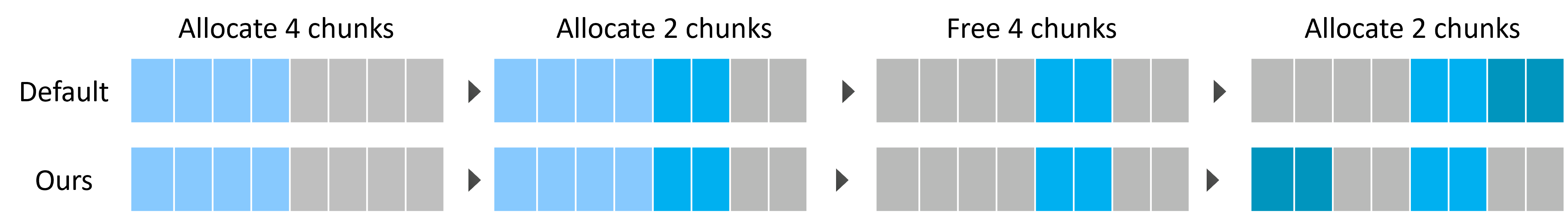}
    \caption{Advanced memory allocator (Ours) vs. default memory allocator (Default)}
    \label{fig:advanced_allocator}
\end{figure}

The default memory allocator usually creates a new buffer each time when receiving a memory allocation request, and therefore the data is less likely to be reused. To maximize the buffer reuse and make the data more cache friendly, we develop an advanced memory allocator as shown in Figure~\ref{fig:advanced_allocator}.

\subsubsection{Weight sharing}
\label{sec:wsharing}
\begin{figure}[htbp]
    \centering
    \includegraphics[scale=0.6]{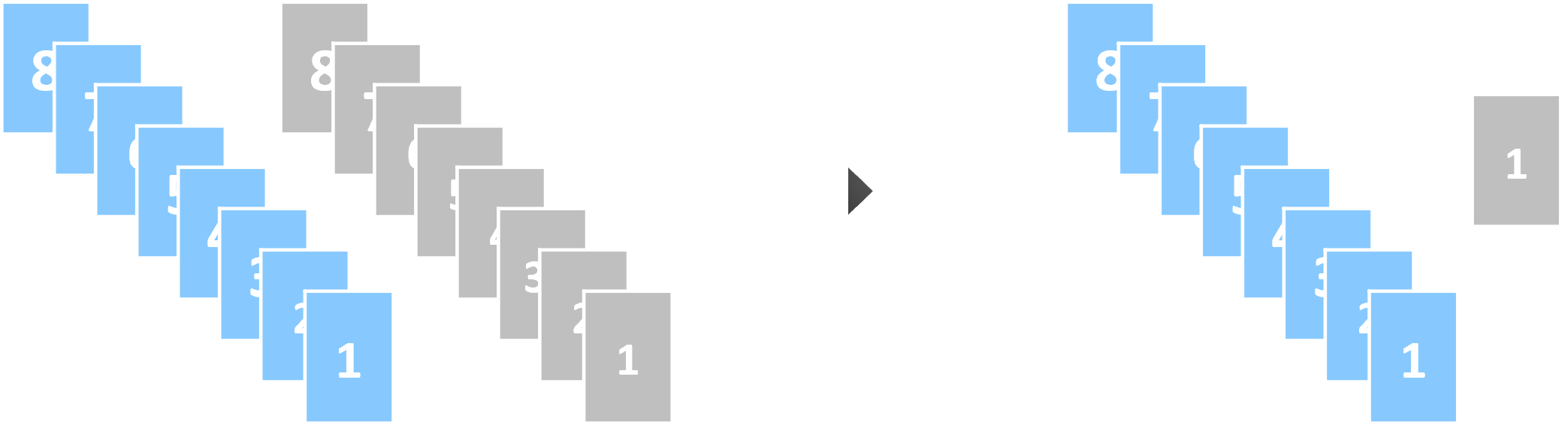}
    \caption{Weight sharing}
    \label{fig:wsharing}
\end{figure}

Weight sharing is another useful runtime optimization used for efficient weight reuse in inference. Figure~\ref{fig:wsharing} shows how weight sharing works. The left one shows each inference instance (blue box) with the separate weight (grey box). If users want to run eight instances in parallel, eight copies of weight are needed, therefore lowering the overall inference efficiency due to lack of memory reuse. The right one shows one shared weight across multiple instances after applying weight sharing mechanism.

\subsubsection{Graph optimization}
\label{sec:gopts}
\begin{figure}[htbp]
	\begin{minipage}[c]{0.33\linewidth}
		\vspace{5pt}
		\centerline{\includegraphics[width=\textwidth]{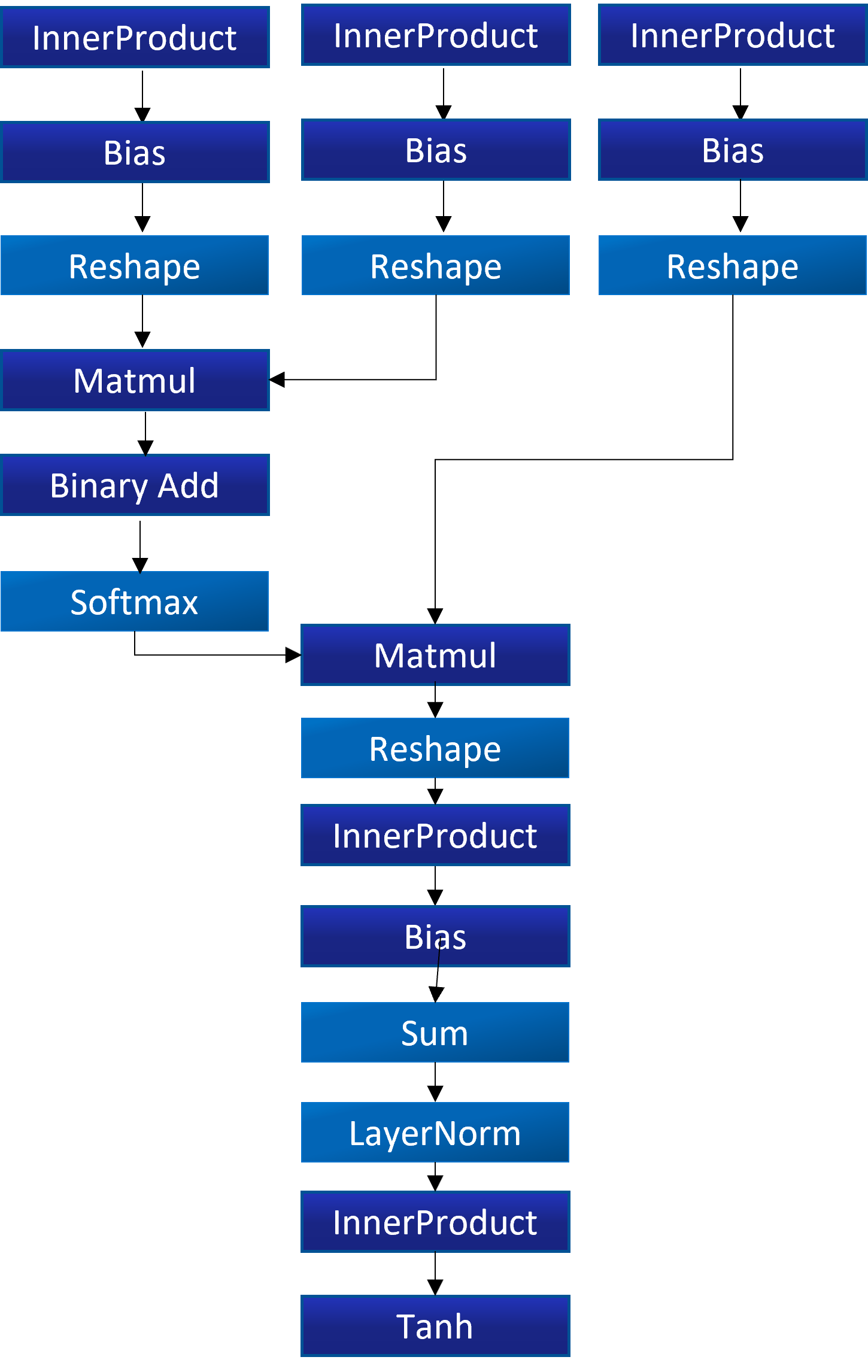}}
		 \centerline{(a)}
	\end{minipage}
	\begin{minipage}[c]{0.33\linewidth}
		\vspace{5pt}
		\centerline{\includegraphics[width=\textwidth]{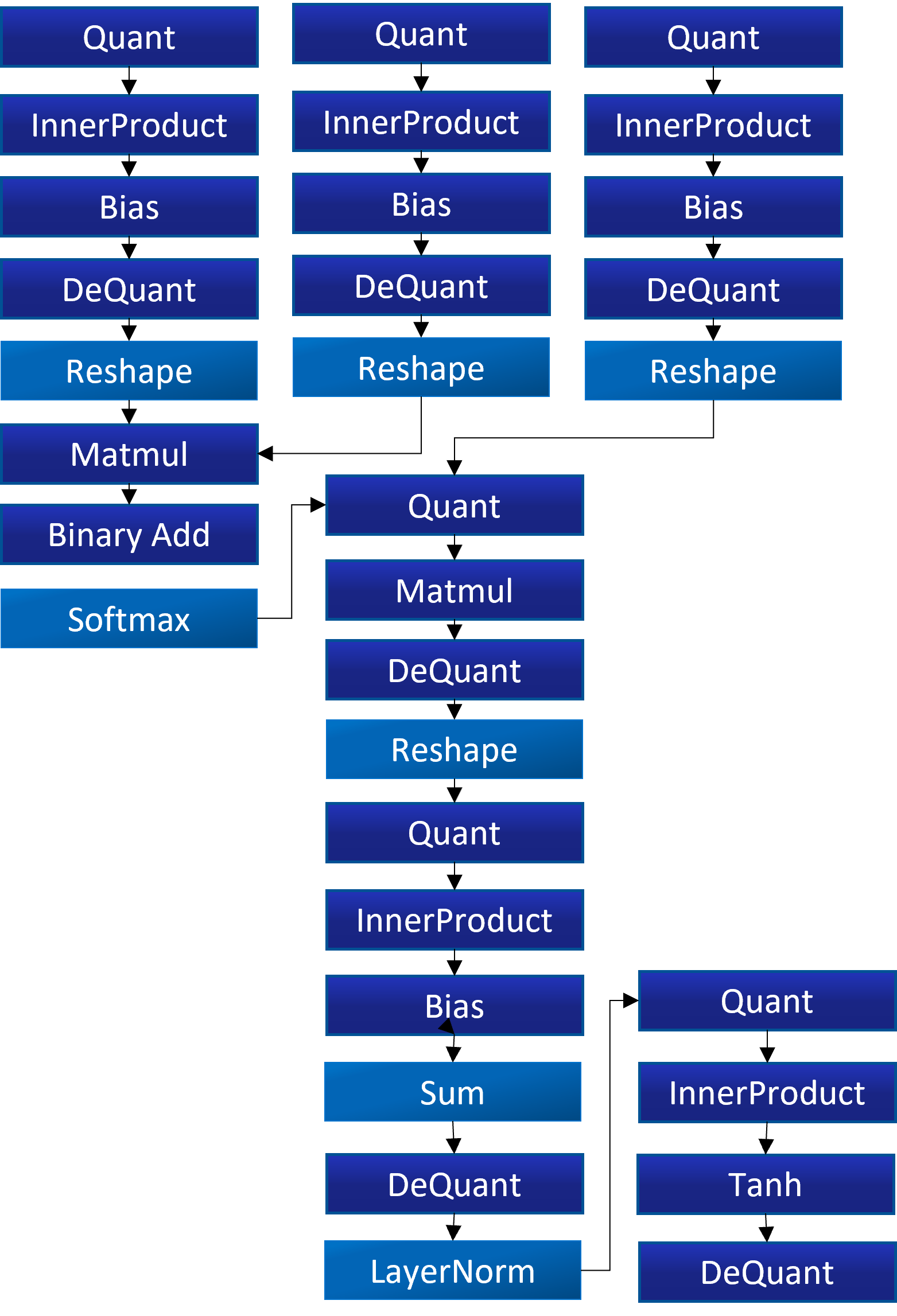}}
		\centerline{(b)}
	\end{minipage}
	\begin{minipage}[c]{0.33\linewidth}
		\vspace{5pt}
		\centerline{\includegraphics[width=\textwidth]{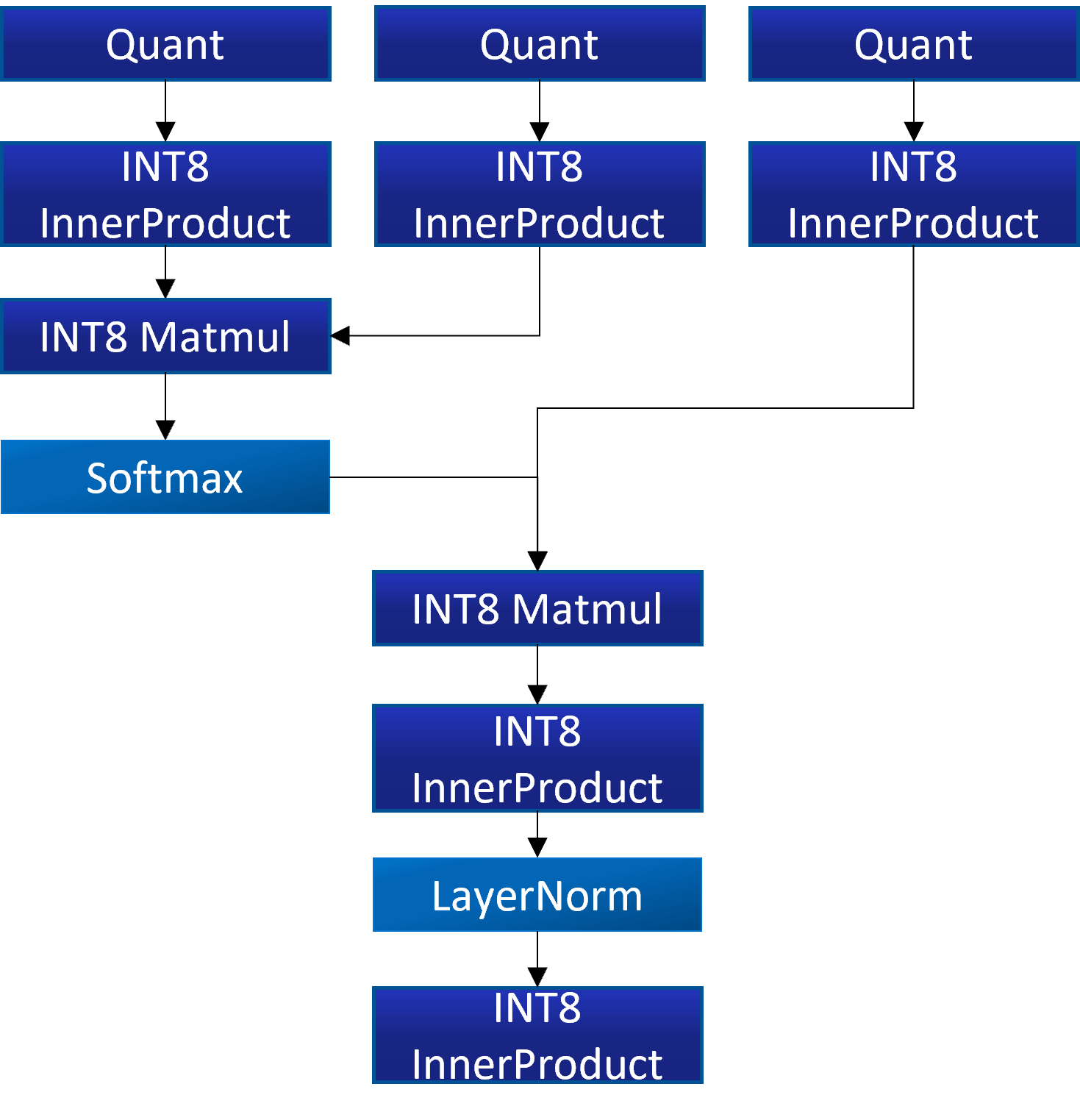}}
		\centerline{(c)}
	\end{minipage}
	\caption{Graph optimization}
	\label{fig:gopts}
\end{figure}

Figure~\ref{fig:gopts} shows the staging transformation from FP32 graph (a), default INT8 graph (b), and optimized INT8 graph (c).
Note that the optimized INT8 graph fuses all the post-operators (e.g., Bias + Reshape, LayerNorm) followed by InnerProduct or Matmul.

\subsubsection{Sparse GEMM operators}
\label{sec:sparse_gemm}
We develop the optimized kernels for sparse GEMM operators based on the pre-defined block-wise structured sparsity pattern with constant block size 4. As described in Section~\ref{fig:sparse}, non-zero weight block is broadcasted to form a VNNI-format block; the sparse weight is compressed and the mask is used to filter the corresponding input as another VNNI-format block for matrix multiplication using VNNI. Algorithm~\ref{algo:sparse} describes the code snippet of INT8 sparse GEMM kernel.

\begin{algorithm}
\label{algo:sparse}
    \caption{Code snippet of INT8 sparse GEMM kernel}
    \begin{algorithmic}
    \STATE // $M, N, K$ as three dimensions of GEMM
    \STATE // $m\_block$ = 4, $n\_block$ = 64, $k\_block$ = 4
    \STATE // $weight\_ptr$ refers to weight tensor, $src\_ptr$ refers to input tensor
    \FOR {$m = 0; m < M; m += m\_block$}
        \FOR {$n = 0; n < N; n += n\_block$}
             \FOR {$k = 0; k <= K; k += k\_block$}
                \STATE $vbroadcastss (\_m32i (weight\_ptr))$
                \STATE $vbroadcastss (\_m32i (weight\_ptr))$
                \STATE $vbroadcastss (\_m32i (weight\_ptr))$
                \STATE $vbroadcastss (\_m32i (weight\_ptr))$
                \FOR {$i = 0; i < 4; ++i$}
                    \STATE $vmovdqu8 (\_m128i, src\_ptr)$
                    \STATE $vmovdqu8 (\_m128i, src\_ptr)$
                    \STATE $vbroadcasti32x4 (\_m512i, \_m128i)$
                    \STATE $vbroadcasti32x4 (\_m512i, \_m128i)$
                    \STATE $vpermt2d (\_m512i, \_m512i, \_m512i)$
                    \STATE $vpshufb (\_m512i, \_m512i, \_m512i)$
                \ENDFOR
                \STATE $vpdpbusd (\_m512i, \_m512i, \_m512i)$
                \STATE $vpdpbusd (\_m512i, \_m512i, \_m512i)$
                \STATE $vpdpbusd (\_m512i, \_m512i, \_m512i)$
                \STATE $vpdpbusd (\_m512i, \_m512i, \_m512i)$
                \STATE // downconvert and post-operator fusion
             \ENDFOR
        \ENDFOR
     \ENDFOR
    \end{algorithmic}
\end{algorithm}

Note that for the weight with 4 non-dividable sparsity dimension, the additional padding is needed while this is not common in NLP models. For simplicity, we omit the special handling of padding in the sparse GEMM kernel implementation.

\subsection{Performance measurement}
\label{sec:model_details}

\subsubsection{Software}
\label{sec:measurement}
We use the latest stable release ONNX Runtime v1.11.1 and Neural Magic v1.1.0 community edition (a436ca67) to measure the performance.

\subsubsection{Performance}
\label{sec:performance}
Table~\ref{throughput_under_latency} shows the absolute maximum throughput (under 10 ms latency constraint) which is used to derive the relative performance in Figure~\ref{fig:perf}.

\begin{table}[htbp]
\caption{Maximum TP under latency constraint on DistilBERT}
\label{throughput_under_latency}
\centering
\begin{tabular}{cccc}
               \toprule
               Sequence Length & \makecell{ONNX Runtime \\(Samples/second)} & \makecell{Neural Magic \\(Samples/second)} & \makecell{Ours \\(Samples/second)} \\
               \midrule
16 &	2713 &	7467 & 11323 \\
32 &	1365 &	4236 &	5440 \\
48 &	930 &	2614 &	3634 \\
64 &	683 &	1990 &	2741 \\
80 &	515 &	1344 &	2010 \\
96 &	437 &	1257 &	1671 \\
112 &	355 &	913 &	1293 \\
128 &	299 &	915 &	1130 \\
               \bottomrule
\end{tabular}
\end{table}

\end{document}